\documentclass[journal]{IEEEtran}
\usepackage{ifpdf}
\usepackage{cite}
\usepackage{algorithmic}
\usepackage{array}
\usepackage{amsmath}
\usepackage{amssymb}
\usepackage{booktabs}
\usepackage{multirow}
\usepackage{dblfloatfix}
\usepackage{url}
\usepackage{float}
\usepackage[table,xcdraw]{xcolor}
\ifCLASSINFOpdf
  \usepackage[pdftex]{graphicx}
\else
  \usepackage[dvips]{graphicx}
\fi

\ifCLASSOPTIONcompsoc
 \usepackage[caption=false,font=normalsize,labelfont=sf,textfont=sf]{subfig}
\else
 \usepackage[caption=false,font=footnotesize]{subfig}
\fi

\ifCLASSOPTIONcaptionsoff
 \usepackage[nomarkers]{endfloat}
\let\MYoriglatexcaption\caption
\renewcommand{\caption}[2][\relax]{\MYoriglatexcaption[#2]{#2}}
\fi

\begin{document}
\title{Revealing the Semantic Selection Gap in DINOv3 through Training-Free Few-Shot Segmentation}
\author{Hussni Mohd Zakir, Eric Tatt Wei Ho \\
Department of Electrical \& Electronics Engineering\\
Universiti Teknologi PETRONAS \\
32610 Seri Iskandar, Perak, Malaysia
}


\maketitle

\begin{abstract}
Recent self-supervised Vision Transformers (ViTs), such as DINOv3, provide rich feature representations for dense vision tasks. This study investigates the intrinsic few-shot semantic segmentation (FSS) capabilities of frozen DINOv3 features through a training-free baseline, FSSDINO, utilizing class-specific prototypes and Gram-matrix refinement. Our results across binary, multi-class, and cross-domain (CDFSS) benchmarks demonstrate that this minimal approach, applied to the final backbone layer, is highly competitive with specialized methods involving complex decoders or test-time adaptation. Crucially, we conduct an Oracle-guided layer analysis, identifying a significant performance gap between the standard last-layer features and globally optimal intermediate representations. We reveal a "Safest vs. Optimal" dilemma: while the Oracle proves higher performance is attainable—matching the results of compute-intensive adaptation methods—current unsupervised and support-guided selection metrics consistently yield lower performance than the last-layer baseline. This characterizes a "Semantic Selection Gap" in Foundation Models: a disconnect where traditional heuristics fail to reliably identify high-fidelity features. Our work establishes the "Last-Layer" as a deceptively strong baseline and provides a rigorous diagnostic of the latent semantic potentials in DINOv3. The code is publicly available at: \url{https://github.com/hussni0997/fssdino}.
\end{abstract}

\IEEEpeerreviewmaketitle
\section{Introduction}

Semantic segmentation models traditionally require large-scale annotated datasets to achieve high performance. However, in niche applications, the cost of obtaining pixel-level annotations is often prohibitive. Few-Shot Semantic Segmentation (FSS) addresses this by enabling models to segment query images using only a handful of labeled support examples.

Despite significant progress, the current FSS landscape faces a persistent challenge: the reliance on supervised source-domain pretraining. Conventional FSS methods often employ complex mask decoders that, while powerful, frequently struggle with domain shift when target classes diverge from the training distribution \cite{Lei2024Cross-DomainSegmentation}. Simultaneously, the rise of foundational models has introduced a new paradigm. While the Segment Anything Model (SAM) and its successors, SAM 2 and SAM 3, offer remarkable class-agnostic segmentation, they are not a "drop-in" solution for class-aware FSS. SAM 3's introduction of concept prompting allows for localized class awareness, but this capability remains largely image-specific and does not inherently bridge the gap to broader categorical generalization across diverse image sets without manual intervention or heavy external pipelines.

Parallel to these developments, self-supervised encoders like DINOv2 and the recently released DINOv3 have demonstrated that large-scale pretraining on diverse datasets yields features with an innate understanding of object boundaries and semantic categories. DINOv3, in particular, utilizes Gram-matrix anchoring and register tokens to stabilize these representations. However, most researchers treat these backbones as "frozen" black boxes, appending trainable segmentation heads that may inadvertently "mask" or "distort" the backbone's intrinsic knowledge through the introduction of new, supervised parameters.

In this paper, we investigate whether the inherent semantic richness of DINOv3 features is sufficient to perform FSS without the need for traditional "induced" learning. We introduce \textbf{FSSDINO}, a training-free baseline that performs segmentation directly using frozen DINOv3 features. By utilizing class-specific prototypes and Gram-matrix refinement, FSSDINO serves as a "pure" probe of the backbone's semantic power. While Test-Time Adaptation (TTA) remains a valuable tool for refining predictions, we argue that a significant portion of the performance gains attributed to TTA may actually be accessible through a simpler, more direct route: optimal feature selection.

Our investigation leads to a startling discovery regarding the distribution of semantic information within Transformer blocks. While the "last layer" is the standard choice for downstream tasks, our \textbf{Oracle-guided analysis} reveals that it is rarely the most semantically performant. We find that the model already possesses the intrinsic capability to reach the performance levels of compute-intensive TTA methods in a single forward pass—provided the optimal intermediate layer is utilized.

However, this leads to a critical \textbf{"Safest vs. Optimal"} dilemma. When attempting to recover this optimal layer using current unsupervised (e.g., Fisher discriminants, register ratios) or support-guided (e.g., support-mIoU, reverse-mIoU) heuristics, these metrics consistently underperform relative to the naive last-layer baseline. This identifies a \textbf{"Semantic Selection Gap"} in foundational vision models: the superior features exist in the intermediate manifold, but they remain "hidden" from current unsupervised selection heuristics.

\noindent\textbf{Our main contributions are summarized as follows:}
\begin{itemize}
    \item We propose \textbf{FSSDINO}, a training-free method for multi-class FSS that achieves competitive results across COCO-$20^i$ and various CDFSS datasets (DeepGlobe, ISIC, SUIM)
    
    \item We demonstrate that DINOv3 features, in their frozen state, are robust to the performance drop-offs typically seen in high N-way multi-class and cross-domain FSS settings.

    \item We provide a rigorous Oracle Study characterizing the "Semantic Selection Gap," proving that while intermediate layers contain SOTA-level information, identifying them via current statistical heuristics is a non-trivial challenge that favors the "safe" but sub-optimal last layer.
    
\end{itemize}

\begin{figure*}
  \centering
   \includegraphics[width=1.0\linewidth]{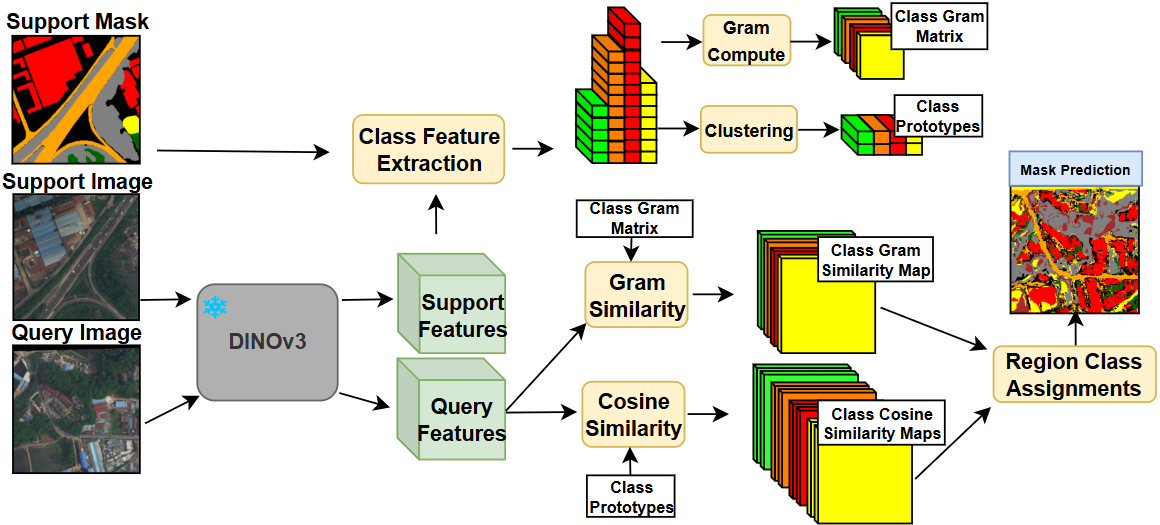}
   \caption{An overview of FSSDINO}
   \label{fig:fssdino}
\end{figure*}

\section{Related Work}
\textbf{Few-Shot Semantic Segmentation (FSS)} aims to segment query images using only a small support set with pixel-level annotations. Early FSS approaches were largely based on meta-learning, particularly Prototypical Networks. PANet \cite{Wang2020PANet:Alignment} introduced prototype alignment regularization to learn a metric space where support and query pixels of the same class are clustered together. PFENet \cite{Tian2022PriorSegmentation} further refined this paradigm by incorporating a training-free prior mask to guide feature enrichment, mitigating bias toward base classes.

Subsequent works increased model complexity by introducing multiple prototypes or more sophisticated feature interactions. ASGNet \cite{Li2021AdaptiveSegmentation} and IPMT \cite{Liu2022IntermediateSegmentation} employed adaptive prototype generation and intermediate feature mining to improve matching quality. These approaches relied heavily on the episodic training protocol, where thousands of simulated few-shot tasks are sampled from base classes. Although effective on benchmarks such as PASCAL-5$^i$, such methods often struggled to generalize to domains with significantly different visual statistics, as the learned matching mechanisms tended to overfit to source-domain textures and geometries.

The introduction of Transformers shifted FSS from global prototype matching to dense pixel-wise correspondence. DCAMA \cite{Shi2022DenseSegmentation} computed full correlation matrices between support and query pixels, enabling fine-grained matching at the cost of quadratic computational complexity. FPTrans \cite{Zhang2022Feature-ProxySegmentation} and CAMSNet \cite{Yan2025CAMSNet:Block} attempted to balance efficiency and accuracy by introducing proxy tokens or hybrid prototype–attention mechanisms. Despite these architectural advances, such methods still relied on episodic meta-training over large base-class datasets.

\textbf{Training-Free FSS with Vision Foundation Models.}
The emergence of vision foundation models (VFMs), particularly DINOv2 \cite{Oquab2023DINOv2:Supervision} and SAM \cite{Kirillov2023SegmentAnything}, enabled a new class of training-free FSS methods. Matcher \cite{Liu2023Matcher:Matching} demonstrated that dense matching of DINOv2 features could rival trained models without parameter updates. PerSAM and PerSAM-F \cite{Zhang2023PersonalizeShot} used support-derived embeddings to prompt SAM’s decoder for one-shot segmentation. These works showed that strong pretrained features can replace task-specific meta-learning.

\textbf{Cross-Domain Few-Shot Segmentation (CDFSS).}
CDFSS evaluates generalization when training and testing domains differ significantly, such as natural images versus medical or satellite imagery. Early approaches, including PATNet \cite{Lei2024Cross-DomainSegmentation} and ABCDFSS \cite{Herzog2024AdaptSegmentation}, attempted to learn domain-invariant prototypes or transformation modules. However, these methods often required unlabeled target-domain data or complex adaptation losses. Recent studies suggest that robust foundation-model features may reduce the need for explicit domain adaptation. FSSDINO supports this perspective by showing that frozen DINOv3 \cite{Simeoni2025DINOv3} features achieve competitive performance on underwater (SUIM) and satellite (DeepGlobe) datasets without domain-specific modules.

\textbf{SAM-Based Auto-Prompting for FSS.}
The Segment Anything Model (SAM) \cite{Kirillov2023SegmentAnything} provides class-agnostic mask generation, making it a central component in many FSS pipelines. However, SAM requires prompts to specify target objects, leading to the problem of automatic prompt generation from support examples. PerSAM \cite{Zhang2023PersonalizeShot} computed a single support embedding as a prompt for SAM. Bridge the Points (GF-SAM) \cite{Zhang2024BridgeSemantically} introduced a graph-based approach that propagates point correspondences between support and query images, followed by SAM-based mask aggregation. SAM-MPA \cite{Xu2024SAM-MPA:Auto-prompting} used deformation fields to propagate masks for medical auto-prompting.

Recently, SAM 3 introduced Promptable Concept Segmentation (PCS), enabling segmentation based on text or visual exemplars. Unlike earlier SAM versions, SAM 3 supports concept-level segmentation using a large internal memory trained on millions of concepts. This shifts the paradigm toward models with intrinsic semantic understanding rather than purely prompt-driven mask generation.

\textbf{Test-Time Adaptation for Few-Shot Segmentation.}
While many recent FSS pipelines rely on strong foundation-model features or promptable segmentation modules, domain shifts in CD-FSS settings can still degrade performance. Test-Time Adaptation (TTA) addresses this issue by adjusting the model to each target episode during inference. Early approaches focused on iterative backbone fine-tuning. For example, IFA \cite{Nie2024Cross-DomainMining} introduces a Bi-directional Few-shot Prediction (BFP) mechanism that establishes correspondences in both support-to-query and query-to-support directions. To mitigate overfitting from limited supervision, it employs an Iterative Few-shot Adaptor that recursively mines these correspondences using augmented support signals.

More recent VFM-based methods instead emphasize parameter efficiency, adapting only small components of otherwise frozen backbones. HERA \cite{HERA:MODELS} proposes a “Select–Regularize–Calibrate” framework that updates fewer than 2.7\% of parameters at test time. It uses Hierarchical Layer Selection (HLS) to dynamically choose the most suitable VFM layer for each episode, followed by entropy-based regularization and pixel-wise calibration. These approaches highlight the growing importance of feature selection and adaptation strategies within frozen foundation models, motivating further investigation into which layers provide the most effective representations for dense prediction.

\textbf{Feature Selection in Vision Transformers.}
An important but underexplored aspect of VFM-based segmentation is the choice of feature layer. Standard practice often uses the final transformer layer, yet recent works such as ALaST \cite{Devoto2024AdaptiveFine-Tuning} and ViT-Split \cite{Li2025ViT-Split:Heads} show that intermediate layers frequently retain richer spatial information for dense prediction tasks.

DINOv3 explicitly addresses the degradation of dense features in large ViTs through Gram Anchoring, a regularization strategy that preserves the spatial correlation structure of patch features. By preventing patch collapse, this mechanism maintains semantic separability across spatial locations. FSSDINO’s analysis of different DINOv3 layers provides empirical insight into the effectiveness of this design for pixel-level tasks.

\textbf{Positioning of This Work.}  
While prior FSS research has primarily focused on designing increasingly complex matching modules, decoders, or adaptation strategies, this work adopts a different perspective. Rather than proposing a new architecture, we investigate the intrinsic few-shot segmentation capabilities of frozen DINOv3 features.

To this end, we introduce FSSDINO as a minimal, training-free diagnostic baseline based on class prototypes and Gram-based refinement. This simple pipeline serves as a controlled probe to evaluate the semantic quality of different feature layers without the confounding effects of meta-training, domain adaptation, or heavy decoders.

Using this setup, we conduct an Oracle-guided layer analysis that reveals a substantial gap between the performance of the conventional last-layer features and the globally optimal intermediate representations. This exposes a previously underexplored “Semantic Selection Gap,” where common unsupervised or support-guided heuristics fail to reliably identify the most semantically discriminative features.

Thus, the primary contribution of this work is not a new segmentation architecture, but a systematic empirical study that characterizes the latent semantic potential of DINOv3 features and establishes the last-layer baseline as both strong and deceptively suboptimal.

\section{Method}

Our proposed framework, illustrated in Fig.~\ref{fig:fssdino}, performs direct semantic segmentation using frozen DINOv3 features. The pipeline consists of two primary stages: \textbf{Class Prototype Construction} and \textbf{Region Class Assignment}, complemented by a \textbf{Gram-based Refinement} mechanism.

\subsection{Class Prototype Construction}

Given a query image $x^q$ and a set of $K$ support images with semantic masks $\mathcal{S} = \{(x^s_k, m^s_k)\}_{k=1}^{K}$, we aim to extract representative class descriptors. Each mask $m^s_k \in \{0, 1\}^{H \times W \times C}$ provides pixel-level annotations for $C$ semantic classes.

All images are processed by a shared frozen DINOv3 encoder, $\text{Enc}(\cdot)$, to extract spatial feature maps:
\begin{equation}
    F^q = \text{Enc}(x^q), \quad F^s_k = \text{Enc}(x^s_k)
\end{equation}
where $F^q, F^s_k \in \mathbb{R}^{h \times w \times d}$. Support masks are bilinearly interpolated to $\tilde{m}^s_k \in \mathbb{R}^{h \times w \times C}$ to match the feature resolution. 

For each class $c \in \{0, \dots, C\}$, 0 being the background, we aggregate all support features corresponding to the class mask across all $K$ support images into a feature set $\mathcal{X}^c = \{f \in \mathbb{R}^d\}$. To obtain a compact representation, we cluster these features into $n_c$ components using $k$-means clustering with cosine distance:
\begin{equation}
    P^c = \text{Cluster}(\mathcal{X}^c, n_c) \in \mathbb{R}^{n_c \times d}
\end{equation}
where $P^c = \{p^c_1, \dots, p^c_{n_c}\}$ denotes the set of \textbf{class prototypes}.

\subsection{Prototype-Based Similarity}

We compute the similarity between the query feature map $F^q$ and the prototypes of class $c$. For each prototype $p^c_i$, a similarity map $S^c_i \in \mathbb{R}^{h \times w}$ is generated:
\begin{equation}
    S^c_i(u,v) = \frac{F^q(u,v) \cdot p^c_i}{\|F^q(u,v)\| \|p^c_i\|}
\end{equation}
This yields a set of $n_c$ similarity maps $\{S^c_1, \dots, S^c_{n_c}\}$ for each class $c$.

\subsection{Gram-Based Refinement}

To complement the first-order alignment of prototypes, we introduce a \textbf{Gram-based refinement} that models inter-channel correlations of support features.

For each class $c$, we reuse the class-specific feature set
\[
\mathcal{X}^c = \{f \in \mathbb{R}^d\}
\]
constructed during prototype extraction, where $|\mathcal{X}^c| = N_c$ denotes the number of support pixels belonging to class $c$.

We first normalize each feature vector:
\begin{equation}
    \tilde{f} = \frac{f}{\|f\|}
\end{equation}

The class-specific Gram matrix is then computed as:
\begin{equation}
    G^c = \frac{1}{N_c} \sum_{\tilde{f} \in \mathcal{X}^c} \tilde{f}\tilde{f}^\top
    \in \mathbb{R}^{d \times d}
\end{equation}

Given the query feature map $F^q \in \mathbb{R}^{h \times w \times d}$, we reshape it into
\[
\mathbf{Q} \in \mathbb{R}^{d \times hw}
\]
and project it using the support Gram matrix:
\begin{equation}
    \hat{\mathbf{Q}} = G^c \mathbf{Q}
\end{equation}

The Gram similarity map is computed as the channel-wise energy between the original and projected features:
\begin{equation}
    S^c_{\text{gram}}(u,v)
    =
    \sum_{j=1}^{d}
    Q_j(u,v)\,
    \hat{Q}_j(u,v)
\end{equation}

The resulting map is normalized to $[0,1]$ and appended to the similarity set:
\[
\mathcal{S}^c = \{S^c_1, \dots, S^c_{n_c}, S^c_{\text{gram}}\}.
\]

\subsection{Region Class Assignment}

All maps in $\mathcal{S}^c$ are bilinearly upsampled to the original resolution $H \times W$. We aggregate these maps using mean and max operations:
\begin{equation}
    \tilde{S}^c_{\text{mean}} = \text{mean}(\mathcal{S}^c), \quad \tilde{S}^c_{\text{max}} = \max(\mathcal{S}^c)
\end{equation}
The final score map for class $c$ is defined as $\tilde{S}^c_{\text{score}} = \tilde{S}^c_{\text{mean}} \odot \tilde{S}^c_{\text{max}}$. Each pixel $(u,v)$ is assigned to the class with the highest score:
\begin{equation}
    \hat{y}(u,v) = \arg\max_{c} \tilde{S}^c_{\text{score}}(u,v)
\end{equation}

\section{Layer-wise Oracle and Heuristic Analysis}

While the final transformer layer is the standard choice for downstream tasks, we observe that intermediate layers in DINOv3 capture varying degrees of semantic and structural information. To systematically identify the optimal features for few-shot segmentation, we perform a layer-wise Oracle analysis followed by an unsupervised heuristic selection study.

\subsection{Oracle Performance and Baseline}

For a transformer with $L$ layers, we extract query features
\[
F^{q,(l)} = \text{Enc}^{(l)}(x^q)
\]
from each layer $l \in \{1, \dots, L\}$ and produce a segmentation prediction $\hat{y}^{(l)}$ using the method described in Sec.~3.

We define two performance bounds:

\begin{itemize}
    \item \textbf{Last-layer Baseline:} The performance achieved using only the final backbone layer $L$.
    
    \item \textbf{Oracle Selection:} The theoretical upper bound where, for each episode $e$, we select the layer $l^*_e$ that achieves the highest mIoU relative to the ground truth $y_e$:
    \begin{equation}
        l^*_e = \arg\max_{l} \text{mIoU}(\hat{y}^{(l)}_e, y_e)
    \end{equation}
\end{itemize}

To ensure a robust global metric, mIoU is calculated by aggregating the confusion matrices of the selected layers across all episodes before computing the mean Intersection over Union.

\subsection{Layer Quality Heuristics}

Let $F^{(l)}$ denote the feature map at layer $l$, and $\hat{y}^{(l)}$ the corresponding query prediction.

\paragraph{Fisher Discriminant Score ($\mathcal{F}$):}
We measure the linear separability of predicted classes in feature space. Let $\mu_c$ be the mean feature vector for class $c$, and $\mu_G$ the global mean. Using cosine distance:
\begin{equation}
    \mathcal{F}^{(l)} =
    \frac{
        \sum_{c} N_c (1 - \cos(\mu_c, \mu_G))
    }{
        \sum_{c} \sum_{f \in \mathcal{X}^c}
        (1 - \cos(f, \mu_c)) + \epsilon
    }
\end{equation}

\paragraph{Reverse mIoU ($\mathcal{M}_{rev}$):}
We derive prototypes from the query prediction $\hat{y}^{(l)}$ and use them to segment the support images. Let $\hat{y}^{s,(l)}_{rev}$ denote the resulting support prediction:
\[
\mathcal{M}_{rev}^{(l)} =
\text{mIoU}(\hat{y}^{s,(l)}_{rev}, y^s).
\]

\paragraph{Support Self-IoU ($\mathcal{M}_{self}$):}
We evaluate how well support-derived prototypes reconstruct the support masks:
\[
\mathcal{M}_{self}^{(l)} =
\text{mIoU}(\hat{y}^{s,(l)}, y^s).
\]

\paragraph{Gram Consistency ($\mathcal{G}$):}
We compute the $L_2$ distance between support and query Gram matrices(derived using the predicted mask):
\begin{equation}
    \mathcal{G}^{(l)} =
    \frac{1}{C}
    \sum_{c}
    \|G^{s,(l)}_c - G^{q,(l)}_c\|_2
\end{equation}

\paragraph{Register-to-Patch Energy Ratio ($\mathcal{R}$):}
Let $r^{(l)}$ and $p^{(l)}$ denote the register and patch tokens:
\[
\mathcal{R}^{(l)} =
\frac{\|r^{(l)}\|_2}{\|p^{(l)}\|_2 + \epsilon}.
\]

\paragraph{Map Entropy ($\mathcal{E}$):}
We measure the sharpness of the similarity maps:
\[
\mathcal{E}^{(l)} =
- \mathbb{E}
\left[
\sum_i p_i \log p_i
\right]
\]
where $p_i$ is the softmax-normalized similarity.

\subsection{Heuristic Optimization via Grid Search}
To find the best proxy for the Oracle, we define a selection score $\mathcal{S}^{(l)}$ as a weighted combination of $M$ heuristics:
\begin{equation}
    \mathcal{S}^{(l)} = \sum_{m=1}^{M} w_m \cdot d_m \cdot \phi(\mathcal{H}_m^{(l)})
\end{equation}
where $d_m \in \{-1, 1\}$ aligns the metric direction and $\phi(\cdot)$ is a transformation function (e.g., $\log(1+x)$ for the Fisher score). We conduct a global grid search over the weight space $\mathcal{W} = \{0.0, 0.1, \dots, 1.0\}$ to identify the configuration that minimizes the gap between $\hat{l} = \arg\max_l \mathcal{S}^{(l)}$ and the Oracle $l^*$.

\section{Experiments}

Our experiments were conducted using the DINOv3 with the ViT-b backbone. For input resolution, we used $512\times512$. We set the parameter $n_c$ to 5 for all runs. Our method was evaluated across various N-way K-shot segmentation scenarios, utilizing mean Intersection-over-Union (mIoU) as the performance metric. In alignment with the MFNET\cite{Zhang2022MFNet:Learning} approach, we excluded the background class from the mIoU calculation and relegated foreground misclassification to be under false positives or false negatives.

To demonstrate the versatility of our method in handling multi-class scenarios across diverse domains, we evaluate its performance on the following datasets: COCO-$20^i$\cite{Shaban2017One-ShotSegmentation}, DeepGlobe\cite{Demir2018DeepGlobeImages}, SUIM\cite{Islam2020SemanticBenchmark}, and ISIC\cite{Codella2019SkinISIC}.

COCO-$20^i$ is a widely-used benchmark for few-shot semantic segmentation (FSS), derived from the MS COCO dataset. It partitions 80 object categories into four non-overlapping folds, each containing 20 classes. For evaluation, we randomly sample 1000 episodes per fold and report the average mean Intersection-over-Union (mIoU) across all four folds for each k-way setting.

DeepGlobe, SUIM, and ISIC serve as cross-domain datasets to further assess generalization. DeepGlobe is a remote sensing dataset with six semantic classes: urban, agriculture, rangeland, forest, water, and barren. SUIM is an underwater scene dataset featuring categories such as divers, plants, wrecks, robots, reefs, fishes, and rocks. ISIC is a medical dataset focused on skin lesion segmentation with three classes. For each of these datasets, we evaluate using 600 randomly sampled episodes per k-way setting to ensure consistency and facilitates fair comparison with existing benchmarks.

\begin{table}[ht]
    \centering
    \caption{Performance comparison on COCO-20$i$ for binary few-shot semantic segmentation.}
    \label{tab:coco20i_binary}
    \begin{tabular}{lcc}
        \toprule
        \textbf{Method} & \textbf{1-shot} & \textbf{5-shot} \\
        \midrule
        PerSAM-F\cite{Zhang2023PersonalizeShot}  & 23.5  & --    \\
        Matcher\cite{Liu2023Matcher:Matching}  & 52.7  & 60.7  \\
        VRP-SAM\cite{Sun2024VRP-SAM:Prompt}  & 53.9  & --    \\
        GF-SAM\cite{Zhang2024BridgeSemantically}   & \textbf{58.7}  & \textbf{66.8}  \\
        DCAMA\cite{Shi2022DenseSegmentation}    & 50.9  & 58.3  \\
        FPTrans\cite{Zhang2022Feature-ProxySegmentation}  & 42.0  & 53.8  \\
        LA\cite{DeMarinis2024LabelPrompts}       & 43.1  & 45.1  \\
        \midrule
        FSSDINO  & 46.99 & 58.54 \\
        \bottomrule
    \end{tabular}
\end{table}

\begin{table*}[ht]
    \centering
    \caption{Comparison on the CDFSS benchmark across different datasets.}
    \label{tab:cdfss_binary}
    \begin{tabular}{lcccccc}
        \toprule
        \multirow{2}{*}{\textbf{Method}} 
        & \multicolumn{2}{c}{\textbf{SUIM}} 
        & \multicolumn{2}{c}{\textbf{DeepGlobe}} 
        & \multicolumn{2}{c}{\textbf{ISIC}} \\
        \cmidrule(lr){2-3} \cmidrule(lr){4-5} \cmidrule(lr){6-7}
        & \textbf{1-shot} & \textbf{5-shot}
        & \textbf{1-shot} & \textbf{5-shot}
        & \textbf{1-shot} & \textbf{5-shot} \\
        \midrule
        GF-SAM \cite{Zhang2024BridgeSemantically}     & --    & --    & 49.5 & 57.7 & 48.7 & 55.2 \\
        PATNet\cite{Lei2024Cross-DomainSegmentation}           & 32.1  & 40.2  & 35.4 & 41.6 & 43.4 & 51.8 \\
        ABCDFSS\cite{Herzog2024AdaptSegmentation}          & 35.1  & 41.3  & 42.6 & 49.0 & 45.7 & 53.3 \\
        IFA\cite{Nie2024Cross-DomainMining}               & --    & --    & 50.6 & 58.8 & 66.3 & 69.8 \\
        HERA$_{\text{DINOv3}}$\cite{HERA:MODELS} 
                          & --    & --    & 44.6 & 63.4 & 61.2 & 73.6 \\
        \midrule
        \textbf{FSSDINO} & \textbf{53.78} & \textbf{62.33}
                          & \textbf{49.41} & \textbf{59.78}
                          & \textbf{55.96} & \textbf{61.67} \\
        \bottomrule
    \end{tabular}
\end{table*}
 
\section{Results and Discussion}
In this section, we evaluate FSSDINO across standard, cross-domain, and multi-class few-shot semantic segmentation (FSS) benchmarks. Our goal is not to outperform all existing methods, but to establish that a fully training-free approach operating on frozen DINOv3 features constitutes a reliable and competitive baseline without architectural specialization or test-time optimization.

\subsection{Standard FSS Benchmarks on COCO-20$i$}
Table~\ref{tab:coco20i_binary} reports binary few-shot segmentation results on COCO-20$i$. FSSDINO achieves 46.99 mIoU in the 1-shot setting and 58.54 mIoU in the 5-shot setting, placing it within the performance range of established FSS methods despite its minimal design.

Notably, FSSDINO substantially improves upon early cosine-similarity baselines such as PerSAM-F, indicating that the dense semantic structure of DINOv3 features—augmented by Gram-matrix refinement—provides a far more stable foundation for training-free segmentation. While methods such as GF-SAM and Matcher achieve higher absolute scores, these approaches rely on significantly more complex pipelines, typically involving SAM-based mask decoders or multi-stage prompting mechanisms.

Importantly, FSSDINO exhibits no evident failure mode on this benchmark: it does not collapse in low-shot settings, nor does it exhibit sensitivity to foreground-background imbalance. Its proximity to methods such as DCAMA, which employ dense multi-level pixel correlations and learned adaptation mechanisms, suggests that a substantial portion of FSS performance is already encoded in the frozen DINOv3 representation itself. When FSSDINO performs well, it reflects the intrinsic quality of the underlying features rather than task-specific optimization.

\subsection{Cross-Domain Generalization (CD-FSS)}
Cross-domain FSS provides a stringent test of representation robustness, as category overlap between source and target domains is minimal. Table~\ref{tab:cdfss_binary} summarizes results on SUIM (underwater), DeepGlobe (satellite), and ISIC (medical) datasets.

Across all three domains, FSSDINO demonstrates consistently strong generalization, achieving the best reported performance among training-free methods in both 1-shot and 5-shot settings. In particular, FSSDINO attains 53.78 mIoU on SUIM (1-shot) and 49.41 mIoU on DeepGlobe (1-shot), indicating that frozen DINOv3 features retain semantic coherence even under severe domain shifts.

While adaptation-based methods such as IFA and HERA outperform FSSDINO in certain 5-shot configurations—most notably on ISIC—these gains are achieved through explicit test-time optimization, including iterative refinement, hierarchical layer selection, or feature recalibration. Such techniques incur additional computational cost and inference latency. In contrast, FSSDINO operates with a single forward pass and no parameter updates.

The absence of catastrophic degradation across domains supports our claim that frozen foundation features provide a robust and dependable baseline for FSS. Rather than relying on domain-specific adaptation, FSSDINO benefits from the generality of the pretrained representation, reinforcing our perspective that much of the apparent need for test-time adaptation stems from suboptimal feature utilization rather than insufficient feature capacity.

\begin{table*}[ht]
\centering
\caption{1-shot, N-way segmentation performance (mIoU) on COCO-$20^i$.}
\label{tab:coco20i_multiclass}
\begin{tabular}{lcccccccc}
\toprule
\textbf{Method} & 1-way & 2-way & 3-way & 4-way & 5-way & 10-way & 15-way & 20-way \\
\midrule
DCAMA\cite{Shi2022DenseSegmentation}  & \textbf{50.9} & 31.7 & 24.2 & 21.1 & 16.7 & 10.8 & 6.5  & 4.7 \\
LA\cite{DeMarinis2024LabelPrompts}     & 43.1 & 34.6 & 31.7 & 29.6 & 27.7 & 23.6 & 16.9 & 13.7 \\
FSSDINO & 47.0 & \textbf{43.7} & \textbf{43.6} & \textbf{44.4} & \textbf{42.6} & \textbf{40.2} & \textbf{38.3} & \textbf{37.4} \\
\bottomrule
\end{tabular}
\end{table*}

\subsection{Robustness in Multi-Class $N$-Way Settings}
Standard FSS benchmarks primarily evaluate binary segmentation, implicitly assuming a single foreground class per episode. To assess scalability beyond this assumption, we evaluate 1-shot $N$-way segmentation on COCO-$20^i$, increasing the number of candidate classes from 1 to 20 (Table~\ref{tab:coco20i_multiclass}).

Across increasing task complexity, FSSDINO exhibits markedly slower performance degradation compared to prior methods. While DCAMA and LA experience steep declines as $N$ grows—dropping to 4.7 mIoU and 13.7 mIoU respectively at 20-way—FSSDINO maintains 37.4 mIoU under the same setting. Performance remains relatively stable between 1-way and 5-way tasks, with only modest decay and occasional local improvements.

These results suggest that frozen DINOv3 features remain highly discriminative even when multiple semantic concepts compete simultaneously. However, we emphasize that this single benchmark alone is insufficient to claim universal superiority of frozen features in multi-class FSS. Instead, it provides strong evidence that prototype-based classification on a high-quality foundation model scales more gracefully than methods implicitly optimized for binary segmentation.

Overall, the $N$-way analysis highlights an important property of FSSDINO: its performance does not hinge on restrictive task assumptions. By avoiding learned decoders and binary-specific inductive biases, the method offers a stable and extensible framework for more complex segmentation scenarios.

\begin{table*}[t]
    \centering
    \caption{Oracle and statistical guidance for feature selection across datasets and shot settings (mIoU \%).}
    \label{tab:oracle}
    \begin{tabular}{lcccccc}
        \toprule
        \textbf{Selection Strategy} 
        & \textbf{DeepGlobe} & \textbf{DeepGlobe}
        & \textbf{ISIC} & \textbf{ISIC}
        & \textbf{COCO-$20^i$ (fold-0)} & \textbf{COCO-$20^i$ (fold-0)} \\
        & \textbf{(1-shot)} & \textbf{(5-shot)}
        & \textbf{(1-shot)} & \textbf{(5-shot)}
        & \textbf{(1-shot)} & \textbf{(5-shot)} \\
        \midrule
        \textbf{Oracle (GT-guided upper bound)} 
        & 59.87 & 65.74 & 66.74 & 75.14 & 53.44 & 64.07 \\
        
        \textbf{Last-layer baseline} 
        & 49.41 & 59.78 & 55.96 & 61.67 & 45.99 & 58.67 \\
        \midrule
        \textbf{Top-weighted heuristic combination} 
        & 45.80 & 57.68 & 48.34 & 59.81 & 39.23 & 58.29 \\
        \midrule
        \textbf{Individual heuristic selection criteria} \\
        \quad Entropy 
        & 40.12 & 43.99 & 34.69 & 40.16 & 10.71 & 13.45 \\
        \quad Regularized Patch-Norm Ratio 
        & 39.30 & 45.74 & 40.73 & 48.61 & 13.65 & 17.71 \\
        \quad Query Prediction Fisher Information 
        & 44.18 & 51.30 & 45.63 & 51.17 & 21.95 & 23.79 \\
        \quad Support Prediction mIoU 
        & 43.60 & 56.56 & 42.85 & 58.55 & 32.30 & 56.92 \\
        \quad Reverse mIoU Consistency 
        & 43.05 & 53.56 & 43.18 & 56.64 & 38.78 & 54.64 \\
        \quad Gram Matrix Distance 
        & 40.43 & 44.65 & 37.80 & 44.04 & 15.60 & 17.87 \\
        \bottomrule
    \end{tabular}
\end{table*}

\subsection{Oracle Analysis and the Semantic Selection Gap}

To probe the latent semantic capacity of the DINOv3 backbone, we conduct an Oracle-guided study in which the optimal feature layer is selected using ground-truth masks. This experiment is not intended as a practical method, but as a diagnostic tool to measure the performance ceiling available within the frozen backbone. The results, shown in Table~\ref{tab:oracle}, reveal a substantial gap between standard layer choices and the globally optimal intermediate representations.

\subsubsection{Latent Potential in Intermediate Layers}

Across all datasets and shot settings, the Oracle-selected layers consistently outperform the last-layer baseline. The gains range from roughly 6 mIoU points on COCO (5-shot) to over 13 points on ISIC (5-shot). This consistent improvement across diverse domains indicates that semantically stronger representations are already present within the intermediate hierarchy of DINOv3, but are not accessed by standard layer selection practices.

The comparison with HERA further clarifies this observation. In the 5-shot setting, HERA reaches 73.6 mIoU on ISIC and 63.4 mIoU on DeepGlobe, with its ISIC result falling within 1.54\% of the Oracle’s theoretical peak (75.14). This proximity suggests that HERA’s primary contribution is not the introduction of fundamentally new semantic information, but rather its ability to identify and exploit the most suitable layer through its Hierarchical Layer Selection (HLS) mechanism.

From the perspective of our Oracle analysis, HERA can be interpreted as a practical strategy that partially closes the \textit{Semantic Selection Gap}. While the Oracle has access to ground-truth masks, HERA relies on support-driven signals to approximate this guidance. The results indicate that such adaptive selection mechanisms can approach the representational ceiling of the frozen backbone, particularly when multiple support examples are available.

This observation reinforces a central finding of our study: the performance ceiling of training-free segmentation is largely determined by layer selection rather than feature quality. The Oracle results demonstrate that substantial semantic capacity already exists within intermediate representations, and that the primary challenge lies in reliably identifying these layers without access to ground-truth supervision.

\subsubsection{Failure of Heuristic Layer Selection}
The central observation in Table~\ref{tab:oracle} is that practical selection strategies fail to recover these optimal layers. Both individual heuristics and their weighted combinations consistently underperform the naive last-layer baseline.

For example, on DeepGlobe 1-shot, the top-weighted heuristic combination achieves 45.80 mIoU, compared to 49.41 mIoU for the last-layer baseline. Similar patterns appear across all datasets: the act of searching for a “better” layer using unsupervised or support-guided criteria often results in worse performance than simply using the final layer.

We refer to this phenomenon as \textit{selection regret}: attempts to optimize layer choice without ground-truth guidance frequently select intermediate representations that appear favorable under local metrics, but are less semantically stable at the mask level. Among the evaluated criteria, support-based measures such as Reverse mIoU Consistency and Support Prediction mIoU are relatively more reliable, particularly in the 5-shot setting. However, even these signals remain substantially below the Oracle upper bound.

\subsubsection{The Last-Layer as a Reliable Default}
The results suggest that the last-layer baseline acts as a reliable default choice rather than a globally optimal one. While intermediate layers can provide higher performance, they also exhibit greater variability and sensitivity to selection criteria. In contrast, the final layer—processed through the full depth of the Transformer—offers a more stable and semantically consistent representation across episodes.

This leads to what we describe as a \textit{“Safest vs. Optimal”} dilemma: the Oracle demonstrates that better-performing layers exist, yet available unsupervised or support-guided metrics are unable to identify them reliably. As a result, the safest practical strategy remains the naive last-layer selection, even though it is not the optimal one.

\subsubsection{The Semantic Selection Gap}
We term this discrepancy the \textit{Semantic Selection Gap}: a disconnect between the semantic potential present within the backbone and the ability of current heuristics to retrieve it. The gap is consistently visible across datasets and shot settings, and persists even when combining multiple selection metrics.

This finding reframes the role of adaptation-heavy FSS methods. Rather than compensating for weak features, many such approaches may be implicitly performing a form of layer or representation selection through iterative refinement. The Oracle results suggest that comparable performance may already be accessible within the frozen backbone, provided that reliable selection mechanisms can be developed.

Consequently, a promising direction for future work lies not in increasingly complex decoders or test-time optimization loops, but in the design of principled, training-free metrics that can bridge this Semantic Selection Gap and consistently identify high-fidelity intermediate representations.

\section{Ablation}

To better understand the factors influencing FSSDINO, we conduct ablation studies on its key components, input resolution, backbone scale, and layer-wise behavior.

\subsection{Influence of Gram-Matrix Refinement}
\begin{table*}[ht]
    \centering
    \caption{Component ablation of FSSDINO, comparing prototype-based matching, Gram-matrix refinement, and their combination across datasets and shot settings (mIoU \%).}
    \label{tab:gram-ablation}
    \begin{tabular}{lcccccc}
        \toprule
        \textbf{Configuration} 
        & \textbf{DeepGlobe} & \textbf{DeepGlobe}
        & \textbf{ISIC} & \textbf{ISIC}
        & \textbf{COCO-$20^i$ (fold-0)} & \textbf{COCO-$20^i$ (fold-0)} \\
        & \textbf{(1-shot)} & \textbf{(5-shot)}
        & \textbf{(1-shot)} & \textbf{(5-shot)}
        & \textbf{(1-shot)} & \textbf{(5-shot)} \\
        \midrule
        Prototype only
        & 48.36 & 58.69 & 53.47 & 59.41 & 44.06 & 58.35 \\
        
        Gram only 
        & 45.92 & 52.60 & 52.77 & 56.55 & 27.41 & 33.38 \\
        
        \textbf{Prototype + Gram} 
        & \textbf{49.41} & \textbf{59.78}
        & \textbf{55.96} & \textbf{61.67}
        & \textbf{45.99} & \textbf{58.67} \\
        \bottomrule
    \end{tabular}
\end{table*}

Table~\ref{tab:gram-ablation} compares three configurations: prototype-only matching, Gram-matrix-only matching, and their combination. Prototype-only matching already provides a strong baseline across all datasets. However, incorporating Gram-matrix refinement yields consistent improvements, particularly in 1-shot cross-domain scenarios. For example, performance on ISIC increases from 53.47 to 55.96 mIoU.

This suggests that the two components play complementary roles. Prototypes capture global semantic alignment between support and query, while the Gram matrix introduces a form of structural or style-based regularization that helps stabilize mask predictions. The significantly lower performance of the Gram-only configuration confirms that structural similarity alone is insufficient for class-aware segmentation and must be anchored by semantic prototypes.
\subsection{Impact of Input Image Resolution}
\begin{table}[ht]
    \centering
    \caption{Effect of input resolution on 1-shot FSSDINO performance across datasets. Results highlight dataset-dependent optimal scales (mIoU \%).}
    \label{tab:imgsz_ablation}
    \begin{tabular}{lccc}
        \toprule
        \textbf{Image Size} 
        & \textbf{DeepGlobe} 
        & \textbf{ISIC} 
        & \textbf{COCO-$20^i$ (fold-0)} \\
        \midrule
        $1024 \times 1024$ & 46.97 & 52.78 & \textbf{48.22} \\
        $512 \times 512$   & 49.41 & \textbf{55.96} & 45.99 \\
        $256 \times 256$   & \textbf{50.15} & 51.73 & 36.41 \\
        \bottomrule
    \end{tabular}
\end{table}

Table~\ref{tab:imgsz_ablation} examines the effect of input resolution on 1-shot performance. The optimal resolution varies across datasets, indicating that training-free segmentation is sensitive to the spatial scale of the underlying visual structures.

On DeepGlobe, performance peaks at the lowest resolution ($256 \times 256$), suggesting that large-scale geographic patterns benefit from broader receptive fields and more globally aggregated features. In contrast, COCO achieves its best performance at the highest resolution ($1024 \times 1024$), where finer spatial detail helps distinguish smaller or more complex objects. ISIC reaches its peak at an intermediate resolution ($512 \times 512$), reflecting a balance between global lesion structure and local boundary detail.

These results indicate that there is no universally optimal input resolution for training-free FSS. Instead, the effective scale depends on the visual characteristics of the dataset and how they interact with the fixed patch structure of the ViT backbone.
\subsection{Effect of Backbone Scaling}
\begin{table}[ht]
    \centering
    \caption{Effect of DINOv3 backbone scale on 1-shot FSSDINO performance. Larger models do not consistently improve training-free segmentation (mIoU \%).}
    \label{tab:mdsz_ablation}
    \begin{tabular}{lccc}
        \toprule
        \textbf{Model Size} 
        & \textbf{DeepGlobe} 
        & \textbf{ISIC} 
        & \textbf{COCO-$20^i$ (fold-0)} \\
        \midrule
        Small ($b$)        & \textbf{49.41} & \textbf{55.96} & \textbf{45.99} \\
        Large ($l$)        & 46.88 & 55.05 & 44.19 \\
        Extra-large ($h+$) & 44.32 & 49.39 & 36.83 \\
        \bottomrule
    \end{tabular}
\end{table}

Table~\ref{tab:mdsz_ablation} compares different DINOv3 backbone scales. Contrary to expectations, larger backbones do not consistently improve training-free segmentation performance. The base model outperforms the large and extra-large variants across all evaluated datasets. For instance, on COCO, performance decreases from 45.99 mIoU (base) to 36.83 mIoU (extra-large).

This result suggests that increased model capacity does not automatically translate into more usable features for direct similarity-based segmentation. Larger models may produce more abstract or globally regularized representations that are less suited to pixel-level matching without task-specific adaptation. In contrast, the base model appears to retain more directly accessible spatial and semantic structure, which benefits training-free prototype-based inference.

\subsection{Layer-wise Reliability Analysis}
\begin{figure}[t]
    \centering
    \includegraphics[width=0.9\linewidth]{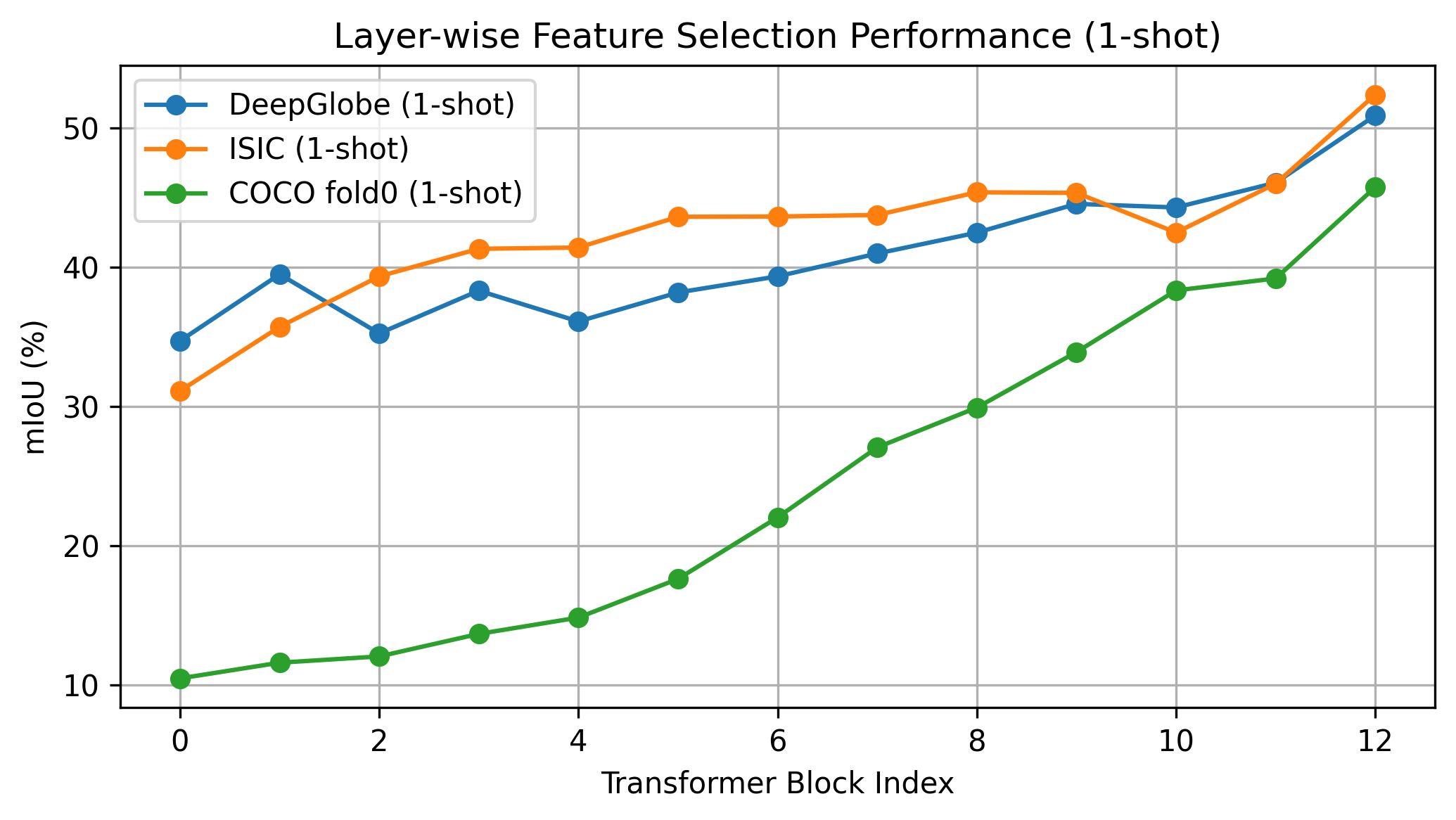}
    \caption{Layer-wise feature selection performance under the 1-shot setting.
    Each curve corresponds to a dataset. Deeper transformer blocks consistently
    yield higher mIoU across datasets on average, indicating that last-layer features provide
    a more reliable selection signal in low-shot scenarios.}
    \label{fig:block_selection}
\end{figure}

We further analyze the reliability of features across the backbone’s depth. As shown in the layer-wise performance curves Fig \ref{fig:block_selection}, the final layer consistently provides the most stable performance across datasets. For example, COCO 1-shot mIoU increases steadily from 10.46 in the earliest block to 45.76 in the final block.

This monotonic trend reinforces the earlier observation from the Oracle study: while intermediate layers may contain higher-performing representations, they are more difficult to identify reliably. The final layer, by contrast, represents the most semantically consolidated and predictable choice. This behavior is consistent with the ``Safest vs. Optimal’’ dilemma: the last layer is not globally optimal, but it remains the most dependable default in the absence of reliable selection criteria.

\subsection{Computational Complexity Analysis}
Since FSSDINO is a training-free framework, its computational cost is dominated by the backbone inference and the analytical operations described in Section 3. To justify the design of our refinement modules, we analyze their theoretical complexity relative to the DINOv3 backbone. Given a feature map of dimension $d$ and $hw$ spatial locations, the complexity of our components is as follows: Prototype Construction: The $k$-means clustering of $N_c$ support pixels into $n_c$ prototypes with $I$ iterations scales as $\mathcal{O}(I \cdot N_c \cdot n_c \cdot d)$. Because $n_c \ll hw$, this aggregation is highly efficient compared to the $\mathcal{O}((hw)^2 \cdot d)$ complexity of the self-attention layers within the frozen encoder.Gram-Matrix Refinement: The construction of the class-specific Gram matrix $G^c \in \mathbb{R}^{d \times d}$ involves a matrix multiplication of size $(d \times N_c) \times (N_c \times d)$, resulting in $\mathcal{O}(N_c \cdot d^2)$. The subsequent projection of the query features $\mathbf{Q}$ scales as $\mathcal{O}(hw \cdot d^2)$. While the Gram-based refinement introduces a quadratic factor with respect to the feature dimension $d$, it avoids the memory-intensive spatial cross-attention mechanisms often used in few-shot segmentation. In practice, $G^c$ is computed only once per support class, adding negligible overhead to the total inference pipeline. Consequently, the additional latency introduced by our clustering and Gram-based refinement is an order of magnitude lower than the backbone's feature extraction, maintaining a throughput nearly identical to that of the base DINOv3 model.

\subsection{Qualitative Results}
\begin{figure*}
  \centering
   \includegraphics[width=\linewidth]{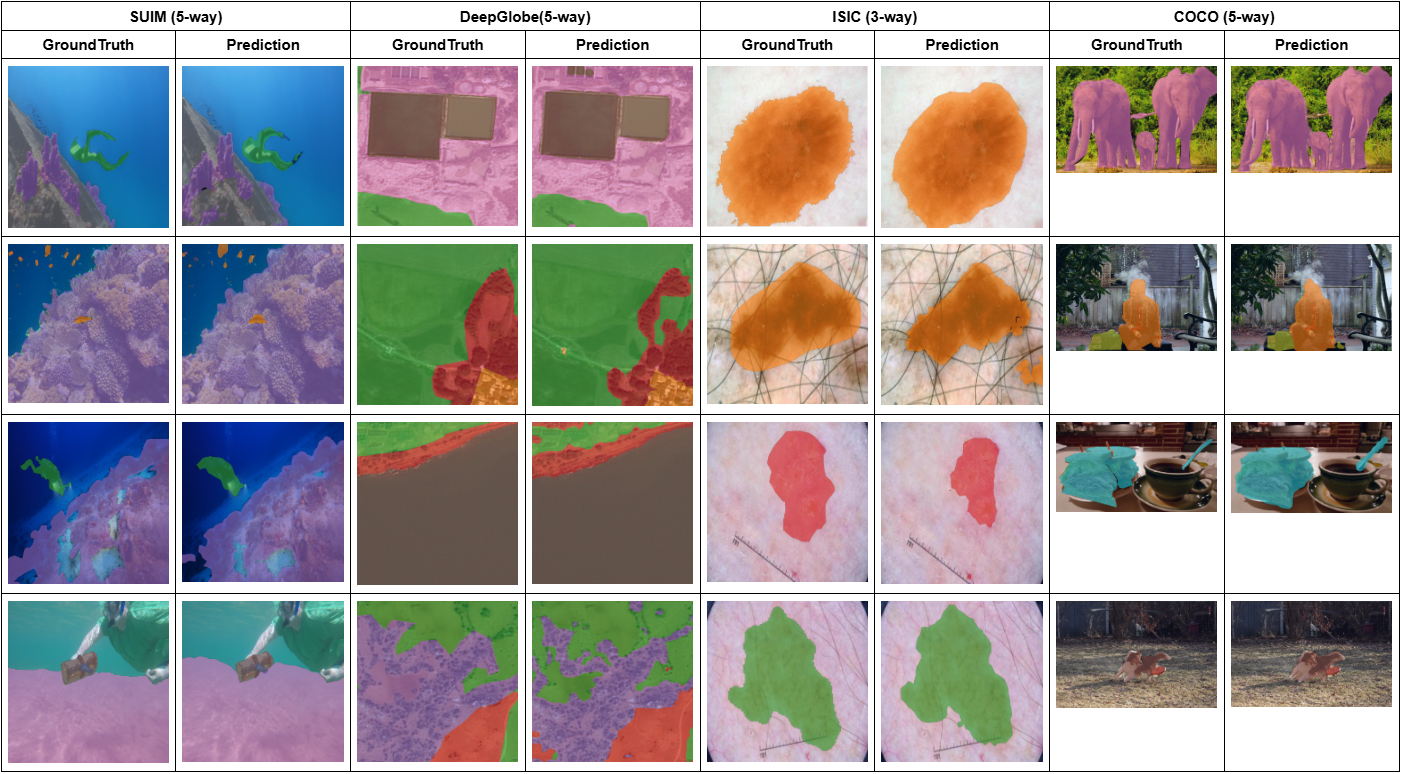}
   \caption{Qualitative segmentation results of FSSDINO on DeepGlobe, ISIC, and SUIM datasets, and MCDINO on COCO-$20^i$.}
   \label{fig:fssdino_vis}
\end{figure*}

In addition, we present qualitative results in Fig.~\ref{fig:fssdino_vis} to further illustrate the effectiveness of our approach. Visualizations from COCO-$20^i$ (5-way), DeepGlobe (5-way), SUIM (5-way), and ISIC (3-way) highlight the model’s ability to accurately segment diverse object classes across varied domains. These examples demonstrate that our method produces consistent and precise segmentations, even in challenging scenarios such as low-contrast underwater scenes in SUIM or heterogeneous lesion appearances in ISIC. The qualitative results complement our quantitative findings and visually affirm the generalization capability of our method across different datasets and class configurations.

\section{Conclusion}

In this work, we investigated the intrinsic few-shot semantic segmentation capabilities of frozen DINOv3 features through a simple, training-free prototype-based framework. Across standard COCO-20$i$ benchmarks, cross-domain CDFSS datasets, and multi-class $N$-way scenarios, our method demonstrates that the last-layer features of a modern self-supervised ViT already constitute a strong and reliable baseline. Despite the absence of learned decoders, test-time adaptation, or architectural specialization, FSSDINO remains competitive with more complex approaches and exhibits no evident failure mode across evaluated settings.

However, our Oracle-guided analysis reveals a substantial gap between the performance of the standard last-layer baseline and the optimal intermediate representations available within the backbone. Across all datasets, Oracle-selected layers consistently yield significant improvements, often approaching the performance of methods that rely on compute-intensive adaptation or specialized decoders. This indicates that the primary limitation of training-free FSS is not the representational capacity of the backbone, but the inability of current heuristics to reliably identify the most semantically faithful feature layers.

We characterize this discrepancy as the \textit{Semantic Selection Gap}: a disconnect between the latent semantic potential of foundation models and the effectiveness of existing unsupervised or support-guided selection strategies. Our experiments show that practical heuristics often perform worse than the naive last-layer choice, creating a “Safest vs. Optimal” dilemma where the most reliable option is not the most powerful one.

Complementary ablation studies further support this perspective. We find that prototype-based matching provides the primary semantic signal, with Gram-matrix refinement offering consistent but secondary improvements. Input resolution proves to be dataset-dependent, and larger backbones do not necessarily yield better training-free segmentation performance. Finally, layer-wise analysis confirms that while intermediate layers can offer higher potential, the final layer remains the most stable default choice.

Overall, our findings suggest that a significant portion of the performance attributed to complex adaptation-based FSS methods may already reside within the frozen backbone. Rather than focusing solely on increasingly sophisticated decoders or optimization loops, future work may benefit more from principled, training-free strategies that can bridge the Semantic Selection Gap and reliably access the high-fidelity representations already present within foundation models.

\ifCLASSOPTIONcaptionsoff
  \newpage
\fi



%


{\small
\bibliographystyle{IEEEtran}

\bibliography{references,references2}
}


%








\end{document}